\documentclass[9pt]{article}
\usepackage{spconf,amsmath,graphicx}
\usepackage{cite}
\usepackage{bibspacing}
\usepackage{booktabs}
\usepackage{color}
\usepackage{amsfonts}
\usepackage{multirow}
\usepackage{amssymb}
\usepackage{array}
\usepackage{authblk}
\usepackage{threeparttable}
\newcommand{\PreserveBackslash}[1]{\let\temp=\\#1\let\\=\temp}
\newcolumntype{C}[1]{>{\PreserveBackslash\centering}p{#1}}
\newcolumntype{R}[1]{>{\PreserveBackslash\raggedleft}p{#1}}
\newcolumntype{L}[1]{>{\PreserveBackslash\raggedright}p{#1}}
\usepackage{enumitem}
\usepackage{subfigure}
\usepackage[hang]{footmisc}
\usepackage{graphicx}
\usepackage[belowskip=-1pt]{caption}
\usepackage[skip=2.0pt]{caption}

\usepackage{lipsum}

\AtBeginDocument{
 \footnotesize\setlength{\footnotemargin}{0pt}\normalsize
 \edef\hangfootparindent{\the\parindent}
}

\makeatletter
\newcommand\email[2][]%
   {\newaffiltrue\let\AB@blk@and\AB@pand
      \if\relax#1\relax\def\AB@note{\AB@thenote}\else\def\AB@note{\relax}%
        \setcounter{Maxaffil}{0}\fi
      \begingroup
        \let\protect\@unexpandable@protect
        \def\thanks{\protect\thanks}\def\footnote{\protect\footnote}%
        \@temptokena=\expandafter{\AB@authors}%
        {\def\\{\protect\\\protect\Affilfont}\xdef\AB@temp{#2}}%
         \xdef\AB@authors{\the\@temptokena\AB@las\AB@au@str
         \protect\\[\affilsep]\protect\Affilfont\AB@temp}%
         \gdef\AB@las{}\gdef\AB@au@str{}%
        {\def\\{, \ignorespaces}\xdef\AB@temp{#2}}%
        \@temptokena=\expandafter{\AB@affillist}%
        \xdef\AB@affillist{\the\@temptokena \AB@affilsep
          \AB@affilnote{}\protect\Affilfont\AB@temp}%
      \endgroup
       \let\AB@affilsep\AB@affilsepx
}
\makeatother

\title{GEmo-CLAP: Gender-Attribute-Enhanced Contrastive Language-Audio Pretraining \\ for Accurate Speech Emotion Recognition}

\author[1]{Yu Pan}
\author[4]{Yanni Hu}
\author[4]{Yuguang Yang}
\author[5]{Wen Fei}
\author[4]{Jixun Yao}
\author[4]{Heng Lu}
\author[2,3$\dagger$]{Lei Ma}
\author[1]{Jianjun Zhao}

\affil[1]{Kyushu University, Japan \quad $^2$University of Tokyo, Japan  \quad $^3$University of Alberta, Canada}
\affil[4]{Ximalaya Inc., ShangHai, China \quad \quad \quad \quad $^5$Shanghai Jiao Tong University, China}

\email{panyu.ztj@gmail.com, \{yanni.hu, yuguang.yang, bear.lu\}@ximalaya.com,  
       fw.key@sjtu.edu.cn, yaoxunji@gmail.com, ma.lei@acm.org, zhao@ait.kyushu-u.ac.jp}

\begin{document}
%\ninept
%
\maketitle
\begin{abstract}
Contrastive cross-modality pretraining has recently exhibited impressive success in diverse fields, whereas there is limited research on their merits in speech emotion recognition (SER).
In this paper, we propose GEmo-CLAP, a kind of gender-attribute-enhanced contrastive language-audio pretraining (CLAP) method for SER.
Specifically, we first construct an effective emotion CLAP (Emo-CLAP) for SER,  using pre-trained text and audio encoders.
Second, given the significance of gender information in SER, two novel multi-task learning based GEmo-CLAP (ML-GEmo-CLAP) and soft label based GEmo-CLAP (SL-GEmo-CLAP) models are further proposed to incorporate gender information of speech signals, forming more reasonable objectives.
Experiments on IEMOCAP indicate that our proposed two GEmo-CLAPs consistently outperform Emo-CLAP with different pre-trained models. 
Remarkably, the proposed WavLM-based SL-GEmo-CLAP obtains the best WAR of 83.16\%, which performs better than state-of-the-art SER methods. 
\end{abstract}


{
\let\thefootnote\relax
\footnote{$\dagger$ denotes the corresponding author.}
}

\vspace{-4mm}
\section{Introduction}
\label{sec:intro}%
\vspace{-2mm}

\noindent
As one of the crucial parts in human-machine interaction, speech emotion recognition (SER) has garnered widespread attention of researchers cross academia and industry \cite{Ando20}.

Over the previous decade, with tremendous progress of deep learning techniques, numerous deep neural network (DNN) based SER methods \cite{LIGHTSERNet,TIMNet} have been proposed and achieved better performance compared with traditional SER methods.
However, the SER field still faces one critical issue which is the lack of sufficient available dataset \cite{Pepino}, since collecting and annotating speech emotion data is expensive and time-consuming.
To this end, several attempts have been made \cite{Chen,Morais,Gat} to utilize self-supervised learning (SSL) based pre-trained models to achieve emotion recognition. 
For instance, chen \emph{et al}. \cite{Chen} introduced three fine-tuning approaches to optimize the Wav2Vec 2.0 \cite{wav2vec2} for SER.
Morais \emph{et al}. \cite{Morais} presented a novel upstream-downstream architecture based modular SER system, which used Hubert \cite{hubert} and Wav2Vec 2.0 based SSL features for emotion recognition.
Gat \emph{et al}. \cite{Gat} proposed a gradient-based adversarial learning framework which normalized the voiceprint information from Hubert features for better recognition performance. 

Despite obtaining impressive performance, these approaches still have several flaws.
First, these methods generally adopt supervised learning based methods to fine-tune pre-trained models, while neglecting research on contrastive learning based approaches.
Second, these methods overlook gender information of speech signals, which has a significant impact on SER.

Hence in this work, we delve into contrastive learning based speech emotion modeling approaches, and introduce GEmo-CLAP which is a kind of gender-attribute-augmented contrastive language-audio pretraining (CLAP) method for SER.
The main contributions are three-fold.
\emph{First}, we build a baseline emotion CLAP (Emo-CLAP) that employs contrastive learning to uncover feature representations across various data pairs. 
As shown in the Fig. \ref{fig:EmoCLAP}, the proposed Emo-CLAP can leverage pre-trained audio and text encoders, enabling the alignment of features from these two modalities into approximately the same feature space.
\emph{Second}, considering the importance of gender information for SER, we further propose a novel multi-task learning based GEmo-CLAP (ML-GEmo-CLAP) and a novel soft label based GEmo-CLAP (SL-GEmo-CLAP), which integrate gender and emotion information of speech signals.
In this way, the proposed two GEmo-CLAPs can form more reasonable objectives, thereby enhancing their recognition performance.
\emph{Third}, we perform experiments on the challenging IEMOCAP \cite{IEMOCAP} corpus to verify the effectiveness of our proposed CLAP-based methods, using one pre-trained text encoder Roberta \cite{roberta}, as well as four pre-trained audio encoders, i.e., Wav2vec 2.0, Hubert, WavLM \cite{wavlm}, and Data2vec \cite{data2vec}. 
Results show that our proposed two GEmo-CLAPs consistently outperform the baseline, while also achieving the best recognition results as opposed to state-of-the-art (SOTA) SER approaches.

To the best of our knowledge, this study marks the first systematic utilization of contrastive cross-modality pretraining to achieve SER. Besides, we aspire for this work to serve as a cornerstone for large-scale generative speech models applicable to text-to-speech synthesis or voice conversion task.

\section{Related Work}
\label{sec:Related Work}
\vspace{-1mm}

\subsection{Speech Emotion Recognition}
\vspace{-1mm}
SER aims to identify the emotional state of different speakers based on their speech signals, which has been widely employed in various practical scenarios \cite{Ando20}.
Although current SOTA DNN-based SER systems \cite{LIGHTSERNet,TIMNet,Pepino,Chen,Morais,Gat} have achieved great performance, there are still several knotty problems need to be solved.
First, from the aspect of data, the collection and annotation of speech emotion data is difficult than the collection of image and text data, which means trained SER models will inevitably encounter unseen data during deployment, thus placing high demands on the overall performance of SER methods.
Second, from the perspective of attribute distribution of speech signals, there are too many mutually influenced speech attributes (emotion, speaker, gender, language, and etc) in human speech \cite{pan2023msac}, which pose great challenges on speech emotion modeling.

\vspace{-1.5mm}
\subsection{Contrastive Cross-modality Pretraining}
\vspace{-1mm}
Contrastive cross-modality pretraining methods have recently been extensively applied in the various fields \cite{CLIP,CLAP,CLAP1,CALM}.
Taking a comprehensive view, all of these methods strive to acquire versatile visual-linguistic or acoustic-linguistic features via contrastive learning.
For example, Radford \emph{et al}. \cite{CLIP} first proposed a novel contrastive language-image pretraining model that learns directly from raw texts about images, which can loosen the restricted supervision of traditional gold-labels. Afterward, Elizalde \emph{et al}. \cite{CLAP} and Wu \emph{et al}. \cite{CLAP1} presented two contrastive learning based language-audio pretraining methods to achieve a unified feature representation of audio and text modalities.
Meng \emph{et al}. \cite{CALM} introduced a contrastive acoustic-linguistic module to capture style-associated text features to improve the speaking style of text-to-speech systems.

\section{METHODOLOGY}
\label{sec:METHODOLOGY}
\vspace{-1mm}

\subsection{Emo-CLAP}
\vspace{-1mm}
Unlike previous studies that employed CLAP for retrieval tasks, this work casts CLAP for the classification task, namely SER. 

\vspace{-1.5mm}
\subsubsection{Training Phase of Emo-CLAP}
\vspace{-1mm}
During training, our proposed Emo-CLAP aims to minimize the distance between paired data within the same class and maximize the distance between data pairs of different classes via contrastive learning, which is depicted in Fig. \ref{fig:EmoCLAP}.
% Unlike previous studies that employed CLAP for retrieval tasks, this work casts CLAP for classification tasks, namely SER. 
% In general, our proposed Emo-CLAP first uses two pre-trained audio and text encoders to process audio-text data pairs and capture their corresponding features, where the text prompt template is \emph{"Emotion is [class]"}. Then, Emo-CLAP projects these cross-modalities' features into approximately the same representation space via contrastive learning, which is depicted in Fig. \ref{fig:EmoCLAP}.

In detail, assume input audio-text data pairs are $\{X_i^a, X_i^t\}$, where $i\in[0, N]$ and N is the batch size. 
During training, Emo-CLAP first extract the text features $F^t$ and audio features $F^a$ via pre-trained audio encoder $f_{a}(\cdot)$ and text encoder $f_{t}(\cdot)$:
\begin{figure}[htbp]
\centering
	\includegraphics[height=6.9cm,width=!]{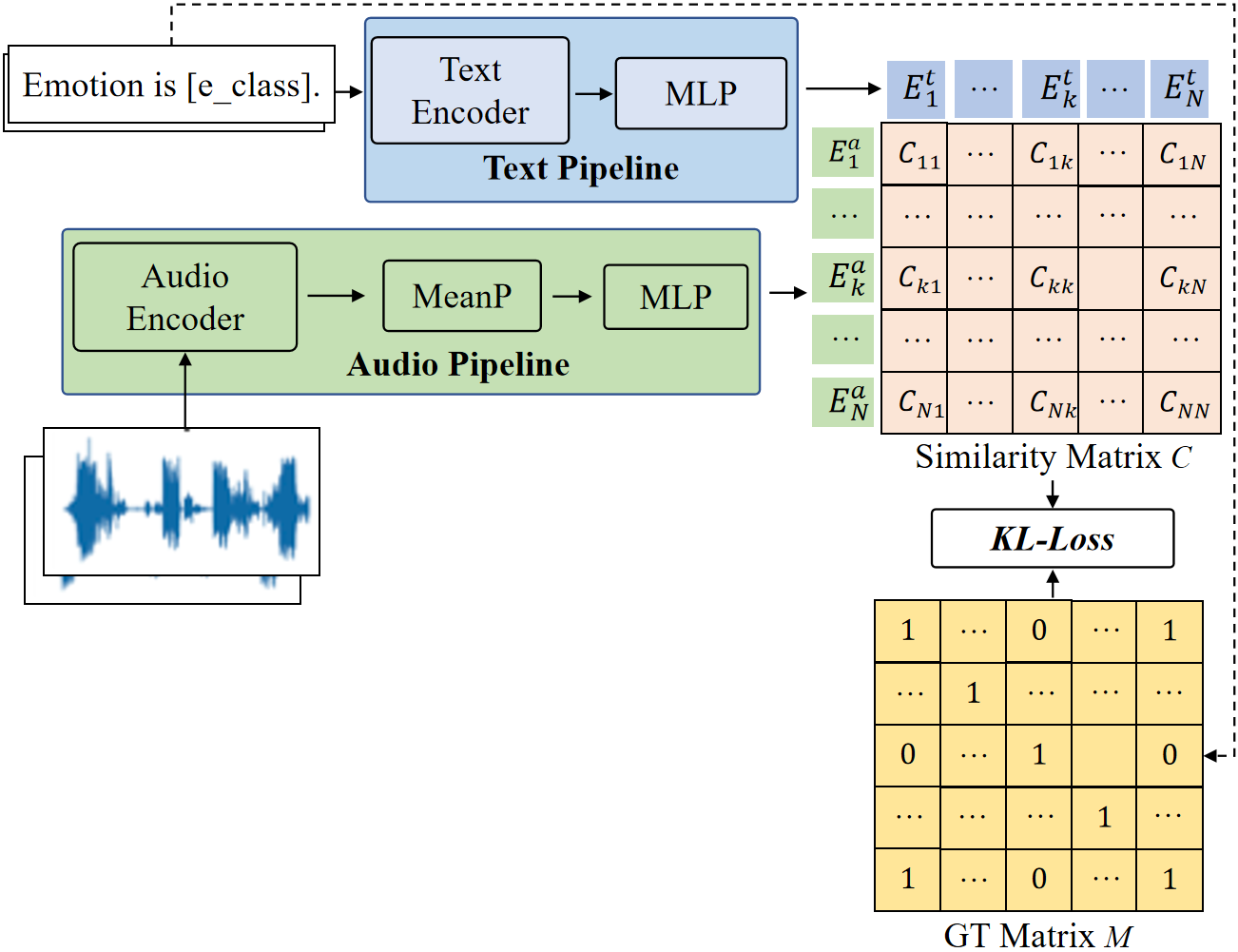}
	\caption{\label{fig:EmoCLAP} Overview of the proposed Emo-CLAP model.}
\end{figure}
\begin{equation}
    \begin{split}
        F^a = f_{a}(X^a);\ \ F^t = f_{t}(X^t);
    \end{split}
\end{equation}
where $F^a\in\mathbb{R}^{N\times T\times D_a}$, $F^t\in\mathbb{R}^{N\times L\times D_t}$, $T$ and $D_a$ are the time length and hidden state dimension of speech signal, $L$ and $D_t$ are the sequence length and hidden state dimension of text.

Next, we apply mean pooling on the temporal dimension of $F^a$ to get processed audio features $\bar{F}^a\in\mathbb{R}^{N\times D_a}$, and use the first element of each sequence of $F^t$ as processed text features $\bar{F}^t\in\mathbb{R}^{N\times D_t}$.
Then, two multi-layer linear projection modules, i.e., $MLP_a(\cdot)$ and $MLP_t(\cdot)$ are designed to bring the processed audio and text features into a joint multi-modal space with the same dimension of $D$, and their corresponding similarity matrices $C_a$ and $C_t$ are measured as:
\begin{equation}
    \begin{split}
        {E}^a = MLP_a(\bar{F}^a);\ \ {E}^t= MLP_t(\bar{F}^t)
    \end{split}
\end{equation}
\begin{equation}
    \begin{split}
        C^a = \varepsilon_a \times ({E}^a \cdot {{E}^t}^T); \ C^t = \varepsilon_t \times ({E}^t \cdot {{E}^a}^T)
    \end{split}
\end{equation}
where ${E}^a\in\mathbb{R}^{N\times D}$ and ${E}^t\in\mathbb{R}^{N\times D}$ denote captured audio and text embeddings, $\varepsilon_a$ and $\varepsilon_t$ are temperature hyper-parameters.

Afterward, Kullback-Leibler divergence based contrastive loss (KL-loss) is used to train Emo-CLAP with the guidance of the inputs' emotional ground truth matrix $M_e\in\mathbb{R}^{N\times N}$. 
It is noteworthy that if true labels between different data pairs are identical within the same batch, their corresponding ground truths are set to 1. Otherwise, their ground truths are set to 0. Therefore, the final total loss of Emo-CLAP is defined as:
\begin{equation}
    \begin{split}
        L_{Total}\!=\!\frac{1}{2}(\!{K\!L}\!({l\_S(C^a)},{S(M_e)}) \!+\! {K\!L}\!({l\_S(C^t)}, {S(M_e)}))
    \end{split}
\end{equation}
\begin{equation}
    \begin{split}
        \text{KL}(P||Q) = \sum_{i,j} P(i,j) \log \frac{P(i,j)}{Q(i,j)}
    \end{split}
\end{equation}
where $S(\cdot)$ denotes softmax function, $l\_S(\cdot)$ denotes log\_softmax function, $i$ and $j$ represent the indexes of the row and column of the two-dimensional matrix.

\subsubsection{Inference Phase of Emo-CLAP}
During inference, we first build natural language supervisions $TC=\{TC_1, ..., TC_N\}$ based on emotion classes $C=\{C_1, ..., C_N\}$.
Then, for a given audio data $\tilde{X_a}$, its category is the best match between its embeddings and natural language supervision embeddings $TC_i$, computed by their cosine similarity.

\subsection{GEmo-CLAP}
Instinctively, the gender information is beneficial for speech emotion modeling, since the speech signals of male and female generally manifest considerable disparities in terms of pitch, tone, energy, and so forth.

Therefore, we further optimize the baseline Emo-CLAP and propose two GEmo-CLAP approaches, i.e., ML-GEmo-CLAP and SL-GEmo-CLAP, which integrate the gender and emotion information of given speech signals, as outlined below in detail.

\subsubsection{ML-GEmo-CLAP}

\begin{figure}[htbp]
\centering
    \setlength{\abovedisplayskip}{-1cm}
	\includegraphics[height=5.0cm,width=!]{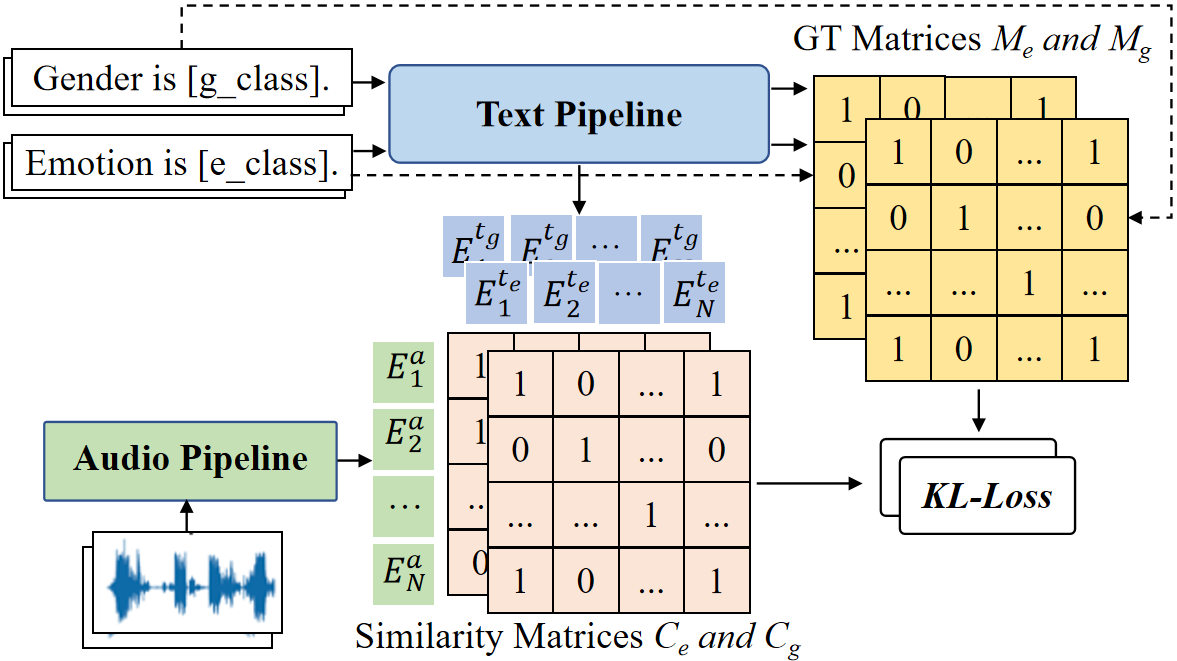}
	\caption{\label{fig:ML-GEmo-CLAP} Overview of the proposed ML-GEmo-CLAP model.}
    \setlength{\belowdisplayskip}{-1cm}
\end{figure}

\noindent
In order to leverage the gender attribute of speech signals, we first propose a novel ML-GEmo-CLAP model, which employs multi-task learning to incorporate gender information into Emo-CLAP. Its overall architecture is illustrated in Fig. \ref{fig:ML-GEmo-CLAP}.

To be specific, our proposed ML-GEmo-CLAP mainly consists of two parts, which are the audio pipeline and shared text pipeline. 
Overall, the proposed ML-GEmo-CLAP is trained via multi-task learning on top of the baseline Emo-CLAP. 
As a result, its final loss $L_{M_{Total}}$ can be defined as:
\begin{equation}
    \begin{split}
        L_{M_{Total}} = \alpha_{e} L_{E_{Total}} + (1-\alpha_{e}) L_{G_{Total}}
    \end{split}
\end{equation}
where $L_{E_{Total}}$ is the KL-loss of emotion attribute, $L_{G_{Total}}$ is the KL-loss of gender attribute, $\alpha_e$ is a parameter to adjust $L_{E_{Total}}$ and $L_{G_{Total}}$. In our case, $\alpha_e$ is basically set to 0.8 or 0.9.

As a consequence, with a more reasonable objective, the proposed ML-GEmo-CLAP model is capable of enhancing its SER performance.

\subsubsection{SL-GEmo-CLAP}

\begin{figure}[htbp]
\centering
    \setlength{\abovedisplayskip}{-8mm}
    \setlength{\belowdisplayskip}{-3mm}
	\includegraphics[height=7.8cm,width=!]{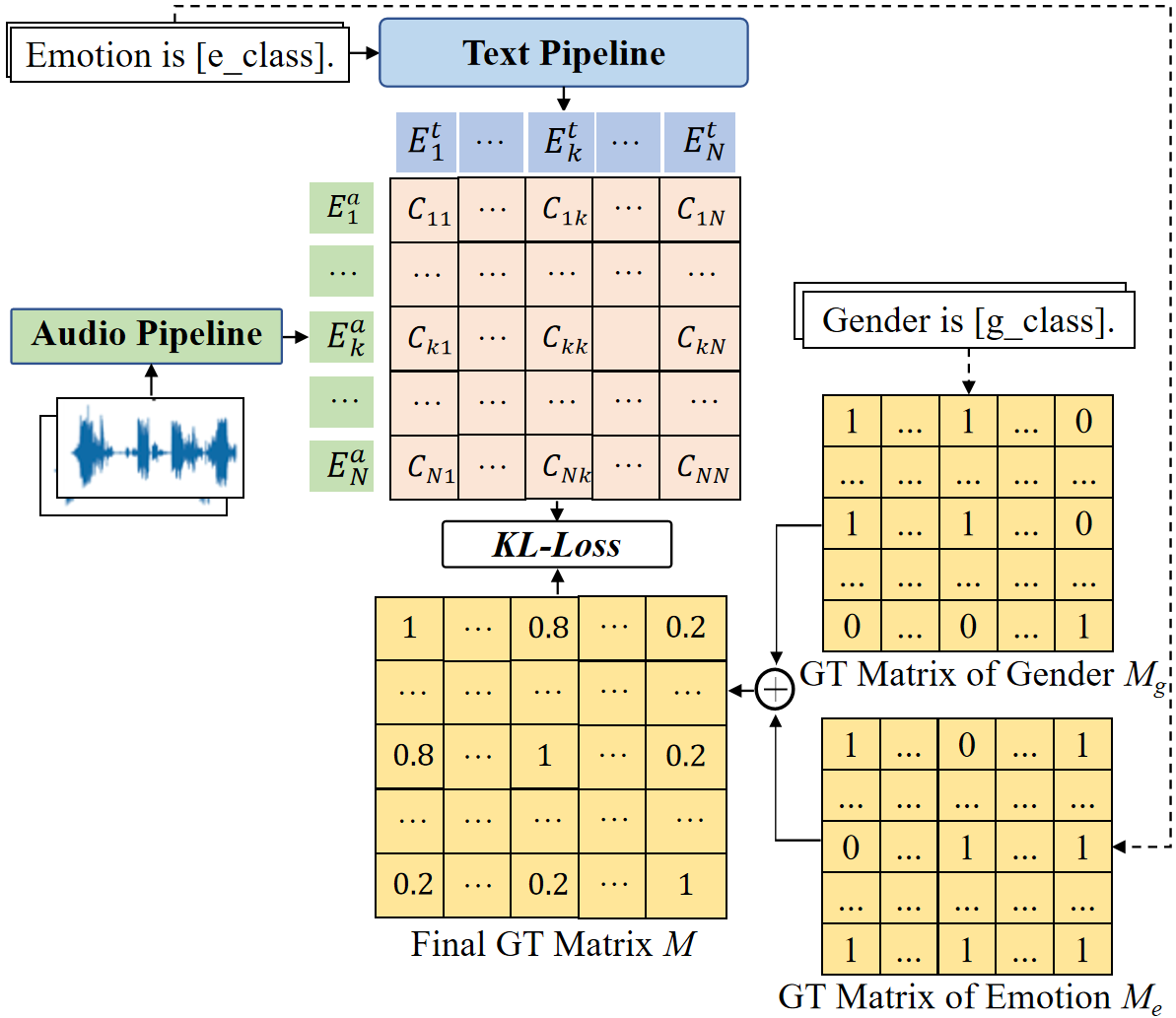}
	\caption{\label{fig:SL-GEmo-CLAP} Overview of the proposed SL-GEmo-CLAP.}
\end{figure}

\noindent
Although ML-GEmo-CLAP can effectively utilize gender and emotion information of human speech, its training requires more computational resources than the baseline Emo-CLAP, which limits its usage to some extent. 
Thus, we further present a novel SL-GEmo-CLAP model based on Emo-CLAP, which leverages gender and emotion attributes of speech signals to create a more reasonable ground truth matrix $M$. 
Fig. \ref{fig:SL-GEmo-CLAP} displays its overall architecture.

More precisely, if the gender categories of different data pairs in the same batch are identical, their corresponding gender ground truth are designated as 1. Conversely, their value are set to 0.
Thus, the final ground truth matrix $M$ is formulated as:
\begin{equation}
    \begin{split}
        M = \alpha_e {M}_e + (1-\alpha_e) {M}_g
    \end{split}
\end{equation}
where $\alpha_e$ is a hyper-parameter to adjust and scale $M_e$ and $M_g$. In our case, $\alpha_e$ is normally set to 0.8 or 0.9.

Accordingly, with a more feasible ground truth matrix, the proposed SL-GEmo-CLAP is able to improve its recognition performance as well.

\section{EXPERIMENTS}
\label{sec:EXPERIMENTS}
\vspace{-1mm}

\subsection{Experimental Settings}
\subsubsection{Databases}
\vspace{-1mm}

As one of the most challenging corpora in SER, IEMOCAP is composed of five sessions, each involving speech from one male and one female speakers. 

In this work, we likewise select four categories (angry, happy+excited, sad, and neutral) which is the same as recent studies to make fair comparisons. 
In addition, a standard 5-fold cross-validation is adopted to evaluate our proposed methods, with one session left out as the test set in each fold.

\begin{table*}[htbp]
    \centering
    \caption{\label{tab:all-results-ours} Overall results of our proposed Emo-CLAP, ML-GEmo-CLAP, and SL-GEmo-CLAP on IEMOCAP.}
    \renewcommand{\arraystretch}{0.65}
    \begin{tabular}{C{35mm}C{20mm}C{35mm}C{32mm}C{32mm}}
        \toprule
         Model    &  Text Encoder  &   Audio Encoder   &   WAR       &  UAR \\
        \midrule
        \multirow{4}{*}{Emo-CLAP}  &  \multirow{4}{*}{Roberta-base}  & Data2vec-large  &  75.86  &  77.71   \\
                                   &                                 &  Wav2vec2-large  &  77.91  &  80.78   \\
                                   &                                 &  Hubert-large    &  \textbf{79.57}  &  80.64   \\
                                   &                                 &  WavLM-large     &  79.28  &  \textbf{81.12}   \\
        \midrule[\heavyrulewidth]
        \multirow{4}{*}{ML-GEmo-CLAP} &  \multirow{4}{*}{Roberta-base}  &  Data2vec-large  &  77.52 (\textbf{\emph{+1.66\%}})  &  80.19 (\textbf{\emph{+2.48\%}})   \\
                                   &                                 &  Wav2vec2-large  &  79.96 (\textbf{\emph{+2.05\%}})  &  \textbf{82.94} (\textbf{\emph{+2.16\%}})   \\
                                   &                                 &  Hubert-large    &  \textbf{80.94} (\textbf{\emph{+1.37\%}})  &  82.30 (\textbf{\emph{+1.66\%}})   \\
                                   &                                 &  WavLM-large     &  \textbf{80.94} (\textbf{\emph{+1.66\%}})  &  82.30 (\textbf{\emph{+1.18\%}})   \\
        \midrule[\heavyrulewidth]
        \multirow{4}{*}{SL-GEmo-CLAP} &  \multirow{4}{*}{Roberta-base}  &  Data2vec-large  &  77.03 (\textbf{\emph{+1.17\%}})  &  79.72 (\textbf{\emph{+2.01\%}})   \\
                                   &                                 &  Wav2vec2-large  &  80.25 (\textbf{\emph{+2.34\%}})  &  81.80 (\textbf{\emph{+1.02\%}})   \\
                                   &                                 &  Hubert-large    &  80.94 (\textbf{\emph{+1.37\%}})  &  82.15 (\textbf{\emph{+1.51\%}})   \\
                                   &                                 &  WavLM-large     &  \textbf{81.43} (\textbf{\emph{+2.15\%}})  &  \textbf{83.16} (\textbf{\emph{+2.04\%}})   \\
        \bottomrule
    \end{tabular}
    \vspace{-2mm}
\end{table*}

\subsubsection{Implementation Details}

In all experiments, we use Adam to optimize our proposed CLAP-based methods with an initial learning rate of 2e-5 and a batch size of 32. All models are trained for 80 epochs within the PyTorch framework.
Besides, for evaluating metrics, we employ weighted average recall (WAR) and unweighted average recall (UAR) to verify the recognition performance of SER methods.

\subsection{Results and Analysis}

\subsubsection{Recognition Comparison of SOTA SER Approaches}
First, we compare the recognition performance of our proposed CLAP-based methods with SOTA SSL-based SER approaches.

\begin{table}[htbp]
    \centering
    \caption{Recognition comparison of SOTA SER methods on IEMOCAP. A and T are audio and text modalities, respectively.}
    \label{tab:comparison-results}
    \renewcommand{\arraystretch}{0.6} 
    \begin{tabular}{C{36mm}C{15mm}C{10mm}C{10mm}}
        \toprule
        Model                      &  Modalities  &   UAR   &  WAR   \\
        \midrule
        Wav2vec2-PT \cite{Pepino}  &  A  &    -    &  67.20  \\
        UDA  \cite{Morais}         &  A  &  77.76  &  77.36  \\
        TAP  \cite{Gat}            &  A  &  74.20  &  -      \\
        KS-Transformer \cite{KS-Transformer} & A + T  &   75.30   &  74.30  \\
        W2B-CA-Aux \cite{sun2023}  &  A + T  &  79.71  &  78.42  \\
        BAM  \cite{Knowledge}      &  A + T  &  77.00  &  75.50  \\
        MMER \cite{ghosh2022mmer}  &  A + T  &    -    &  81.20  \\
        SA-CNN-BLSTM \cite{li2019improved}  &  A  &    82.80    &  81.60  \\
        CNN-CasA-Tri \cite{liu2023discriminative}  &  A  &   82.67   &  \textbf{82.68}  \\
        \midrule
        Emo-CLAP (Ours)            &  A + T  &  81.12  &  79.28  \\
        ML-GEmo-CLAP (Ours)        &  A + T  &  82.30  &  80.94  \\
        SL-GEmo-CLAP (Ours)        &  A + T  &  \textbf{83.16}  &  81.43  \\
        \bottomrule
    \end{tabular}
    \vspace{-2mm}
\end{table}

From Table \ref{tab:comparison-results}, we can observe that all of our proposed methods achieve superior recognition performance compared with SOTA SER methods, which validates the effectiveness of our proposed methods.
Notably, the proposed SL-GEmo-CLAP obtains the best WAR of 81.43\%, which outperforms other SOTA SER approaches.

\subsubsection{Comparison of Our Proposed CLAP-based Approaches}

Then, we compare the performance of the proposed three CLAP-based methods on IEMOCAP, which is illustrated in Table \ref{tab:all-results-ours}. 

It can be easily seen that both SL-GEmo-CLAP and ML-GEmo-CLAP consistently perform better than the baseline Emo-CLAP, which proves the effectiveness and robustness of our gender-attribute-enhanced strategy.
In addition, SL-GEmo-CLAP and ML-GEmo-CLAP exhibit similar performance.
Take WavLM-based models as an example. 
Our Emo-CLAP obtains the the best UAR of 81.12\% and the secondary best WAR of 79.28\%. 
In contrast, the proposed ML-GEmo-CLAP attain the best WAR of 80.94\% and the secondary best UAR of 82.30\%, which performs better than Emo-CLAP by 1.66\% and 1.18\%.
Meanwhile, the proposed SL-GEmo-CLAP achieves the best WAR of 81.43\% and the best UAR of 83.16\%, which outperforms the baseline by 2.15\% and 2.04\%.

\vspace{-1mm}
\subsubsection{Comparison of Different Audio Encoders}
\vspace{-1mm}

Third, we compare and analyze the performance of different pre-trained audio encoders within our proposed CLAP-based SER architecture.

As can be observed in Table \ref{tab:all-results-ours}, when using the pre-trained Roberta-base model as the text encoder, the WavLM-based CLAP approaches generally obtain the best recognition results under all paradigms. 
In addition, Hubert-based CLAP methods achieve the secondary best recognition performance.
Furthermore, Wav2vec2-based CLAP methods normally perform better than Data2vec-based CLAP methods.

\vspace{-0.5mm}
\section{CONCLUSIONS}
\label{sec:CONCLUSIONs}
\vspace{-1mm}
In this study, we introduce GEmo-CLAP, which is a kind of gender-attribute-enhanced contrastive language-audio pretraining method. 
To be precise, a baseline Emo-CLAP model is first built, which is able to leverage pre-trained models from both audio and text modalities. Then, two novel ML-GEmo-CLAP and SL-GEmo-CLAP are further proposed, that incorporate gender information to form more reasonable objectives.
Extensive experiments on IEMOCAP show that our proposed SL-GEmo-CLAP and ML-GEmo-CLAP consistently outperform the baseline Emo-CLAP, while also obtaining the best recognition results as opposed to SOTA SER methods.

\vfill\pagebreak
\label{sec:refs}

\bibliographystyle{ieee}
\bibliography{refs.bib}

\end{document}